# Haptic Shared Control in Steering Operation Based on Cooperative Status Between a Driver and a Driver Assistance System

Ryota Nishimura, Takahiro Wada, and Seiji Sugiyama
Ritsumeikan University

Haptic shared control is expected to achieve a smooth collaboration between humans and automated systems, because haptics facilitate mutual communication. A methodology for sharing a given task is important to achieve effective shared control. Therefore, the appropriate cooperative relationship between a human operator and automated system should be considered.

This paper proposes a methodology to evaluate the cooperative status between the operator and the automated system in the haptic shared control of a steering operation using a pseudo-power pair of torque from each agent and the vehicle lateral velocity as each agent's contribution to vehicle motion. This method allows us to estimate cooperative status based on two axes: the initiative holder and the intent consistency between the two agents. A control method for a lane-keeping assist system (LKAS) that enables drivers to change lanes smoothly is proposed based on the estimated cooperative status. A gain-tuning control method based on the estimated cooperative status is proposed to decrease the assistance system's pseudo-power when intent inconsistency occurs. A method for switching the followed lane to match the driver's and assistance system's intentions is also proposed. A user study using a driving simulator is conducted to demonstrate the effectiveness of the proposed methods. The results demonstrate that the proposed methods facilitate smooth driver-initiated lane changes without significantly affecting the driver's torque or steering wheel angle while significantly improve lane-keeping performance.

## 1. Introduction

Driver assistance systems (DASs) have been developed to reduce driver workload and mitigate or avoid collisions. Emerging highly automated DASs, such as lane-keeping assist systems (LKASs) and adaptive cruise control (ACC), increase the chances that a human operator will collaborate with the DAS. For such collaborative DASs, haptic shared or haptic guidance control has drawn much attention as a control method for the human-machine interface, because it enables the human operator to interact and communicate continuously with the automated system through a haptic interface (see Abbink, Mulder, & Boer, 2012, for an overview). The smooth transfer of control authority is of importance for an effective human-automation system (Inagaki, 2003). Continuous physical interaction via haptic shared control enables a human operator and automated system to smoothly share operational tasks. In automotive safety, the haptic pedal (Mulder, Abbink, & van Paassen, 2011) and steering controls for lane-keeping assistance, as well as curve negotiation (Abbink & Mulder, 2009; Forsyth & MacLean, 2006), have been developed as examples of haptic shared controls. Flemisch et al. (2003) proposed the H-Metaphor as a guideline for achieving smooth vehicle automation interaction. Goodrich and Schutte (2008) applied the H-Metaphor to the interface design of aircrafts.

To analyze the relationship between humans and automation, scales for describing the level of automation (LOA), which characterizes the allocation of authority, have been proposed (Endsley, 1987; Sheridan, Verplank, & Brooks, 1978). The concept of adaptive automation was proposed to appropriately reallocate the functions between humans and machines based on the LOA (Inagaki, 2003; Sheridan, 1992). The LOA and authority of human-machine systems cannot be applied to shared control systems, since they do not address the situation when a human operator and automated system execute an operational task simultaneously. For haptic shared control, Abbink et al. (2012) introduced the concept of the level of haptic authority (LoHA), which is defined as follows: "LoHA constitutes how forceful the human-automation interface connects human inputs and automation inputs and mainly addresses the provided support on a skill-based level through a single control interface." Abbink and Mulder (2009) proposed a method to tune the LoHA using the impedance of a haptic interface around its desired angle. The effect of control parameters or LoHA on driving behavior, including biomechanical responses, was investigated experimentally based on the musculoskeletal model (Abbink & Mulder, 2009; Abbink, Cleij, Mulder, & Van Paassen, 2012). These results suggest that it is effective to tune the LoHA of the haptic shared control according to the cooperative relationship between the human operator and DAS. However, a methodology to determine the desired dimension of a LoHA in a systematic manner has not yet been established because of the difficulty in quantifying the cooperative status.

The purpose of the present paper is to propose a methodology for evaluating the cooperative status between the human operator and the automated system in the haptic shared control of a steering operation. Additionally, a shared control method that adapts to the cooperative status is proposed to achieve smooth cooperative status changes. In shared control systems, the human operator and automated system have their own control intentions; for example, in the lane-keeping task, the human and automated system have different goals in choosing the lane that the vehicle should drive in. Consistency between intentions is important if the two agents are to cooperate. It is also important which agent, either the driver or DAS, holds the initiative in the control. Therefore, the present study proposes a method for judging the cooperative status between the driver and DAS in the context of haptic shared control of steering operations based on two axes: the initiative holder in control and consistency of intentions between the driver and DAS. The pseudo-power and pseudo-work exerted on the vehicle motion by the steering input of a human driver or DAS control actuator are used to analyze the cooperative relationship between these two agents. In addition, a method to guide a cooperative status to a desired one is proposed by adaptively tuning the control gain to the estimated cooperative status for the design of an LKAS that enables the driver to change lanes smoothly. Preliminary versions of this paper appeared in Wada, Nishimura, and Sugiyama (2013) and Nishimura, Wada, and Sugiyama (2013) as a conference proceeding and extended abstract of a conference position paper, respectively. These two papers reported the basic ideas on cooperative status and its application to an LKAS. The present paper is a refined, archival version of these ideas and provides a detailed definition of the cooperative status based on human muscle torque, which is derived using a physical interaction model between the driver and DAS. We also provide experimental results on the driver's torque to show how the proposed method works as well as further discussion on cooperative status and experimental results.

In Section 2, the basic mathematical model of the haptic shared control system for steering operations is presented. In Section 3, the cooperative status between a human operator and the DAS is formulated based on two axes: the initiative holder in control and the consistency of intentions between the driver and DAS. Section 4 describes a control method for the LKAS that allows the driver to change lanes smoothly by adaptively gain-tuning to the estimated cooperative status. Furthermore, a method for resolving the inconsistency of intentions (e.g., changes in the target lane to be followed) is described for the proposed LKAS. Section 5 contains our experimental methods, and results are presented in Section 6. Section 7 discusses these results, and Section 8 contains our conclusions.

## 2. Basic Model of the Haptic Steering System

In this section, we derive a mathematical model for the physical interaction between a human driver and DAS in the shared control. Assume that the driver firmly grasps the steering wheel and rotates it. A motor is attached to the steering shaft. The dynamic equation in one degree of freedom (DOF) motion around the rotation axis of the steering wheel, including the actuator of the DAS, is given by Eq. 1.

$$I_{str}\ddot{\theta} + b_{str}\dot{\theta} = \tau_c + \tau_{das} + \tau_v, \tag{1}$$

where $I_{str}$ is the moment of inertia of the steering mechanism around the rotational axis, $b_{str}$ (scalar) is the damping coefficient of the steering mechanism, $\tau_c$ is the torque exerted by the human hand on the steering wheel at the contact point, $\tau_{das}$ denotes the torque exerted by the actuator of the DAS, $\tau_v$ is the torque exerted on the steering shaft by the vehicle motion, including the self-aligning torque (SAT), and $\theta$ is the steering wheel angle. The dynamic equation of motion of the human driver's arm, simplified to one DOF motion around the steering axis, is given by

$$I_{dr}\ddot{\theta} + b_{dr}\dot{\theta} = \tau_{msl} - \tau_c, \tag{2}$$

where $I_{dr}$ is the moment of inertia of the human arm around the rotational axis, $b_{dr}$ is the damping coefficient of the human arm, and $\tau_{msl}$ is the torque generated by the human arm muscle.

Fig. 1 presents a block diagram of the entire system, where $y$ denotes the lateral position of the vehicle when considering its lateral control. Two agents, the human driver and DAS, cooperatively operate one plant (i.e., the steering mechanism to achieve the desired vehicle motion).

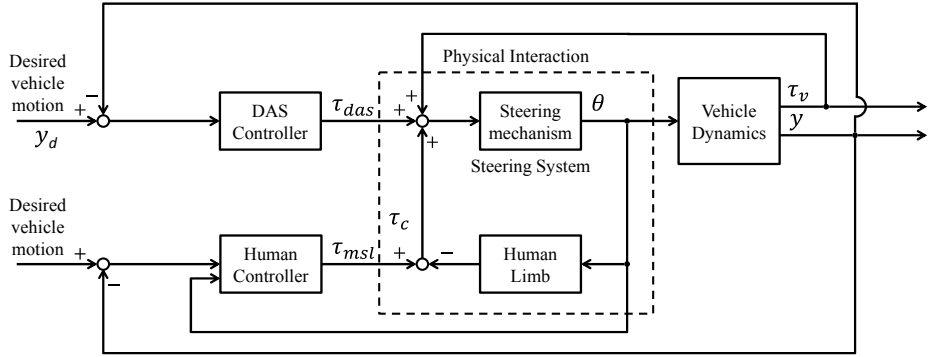

Figure 1. Block diagram of a driver-steering system with a DAS. The human and DAS control the vehicle by sharing the steering operation. When lateral control of the vehicle is considered, $y$ denotes the lateral position of the vehicle on the road. The physical interaction between the human limb and steering mechanism is represented in the middle of the diagram. The human controller determines the muscle torque $\tau_{msl}$ from the feedback information of the vehicle motion and the current steering position $\theta$. The DAS controller outputs the torque $\tau_{das}$ from the feedback information of the vehicle's motion.

## 3. Cooperative States Between the Human and DAS

### 3.1 Pseudo-Power and Work

We investigate the effect of two agents' steering inputs on vehicle motion; specifically, we focus on lateral control of the vehicle. We consider three pseudo-power pairs, $p_c$, $p_{msl}$, and $p_{das}$, defined in Eqs. 3–5, as indices of the influence of each agent's steering input on the vehicle's lateral motion.

$$p_c := \tau_c \dot{y}, \tag{3}$$

$$p_{msl} := \tau_{msl} \dot{y}, \tag{4}$$

$$p_{das} := \tau_{das} \dot{y}, \tag{5}$$

where $\dot{y}$ is the lateral velocity of the vehicle. The pseudo-power takes a positive value when the torque and lateral velocity are in the same direction.

The pseudo-work exerted on the steering system by contact force, muscle torque of the driver, and DAS are as follows:

$$w_c(t) := \frac{1}{\Delta T} \int_{t-\Delta T}^{t} p_c(s) ds, \tag{6}$$

$$w_{msl}(t) := \frac{1}{\Delta T} \int_{t-\Delta T}^{t} p_{msl}(s) ds, \tag{7}$$

$$w_{das}(t) := \frac{1}{\Delta T} \int_{t-\Delta T}^{t} p_{das}(s) ds, \tag{8}$$

where $\Delta T$ is the time window for the work calculation. Note that it is difficult to measure the muscle torque of the driver ($\tau_{msl}$); thus, the pseudo-power, $p_{msl}$, and pseudo-work, $w_{msl}(t)$, were not calculated in this study.

3.2 Cooperative Status Between a Human and Automated System

In this paper, we propose a methodology to evaluate the cooperative status between a human and an automated system based on the following two axes:

a) Initiative holder
b) Intent consistency

a) Initiative holder: The initiative holder is the agent that has greater control of the vehicle motion. Note that $w_c(t)$ is the pseudo-work flowing from the human limb into the steering mechanism through physical interaction (Fig. 1). Therefore, a human driver has initiative when the following is satisfied:

$$w_c(t) \geq -\gamma_1^2, \tag{9}$$

where $\gamma_1^2$ (scalar) is the offset of the judgment threshold; ideally, this scalar should be zero. In this paper, the value is set to $\gamma_1^2 = 0.2$ to prevent incorrect judgments on the negative side due to sensor noise.

b) Intent consistency: The intent consistency is whether the human driver and DAS have the same operations intent. The intent of the two agents is consistent when the following is satisfied:

$$w_{das}(t) \geq -\gamma_2^2 \text{ and } w_c(t) \geq -\gamma_1^2, \tag{10}$$

where $\gamma_2^2$ (scalar) is the offset of the judgment threshold; ideally, this scalar should be zero. Herein, we fix $\gamma_2^2 = 0.1$ to prevent incorrect judgments on the negative side due to sensor noise. The intent of the two agents is defined to be inconsistent when the inequalities point in different directions.

The cooperative status of the two agents is defined in Table 1 according to the initiative holder and intent consistency using $w_c(t)$ and $w_{das}(t)$.

State I: Driver-led cooperative state
    The driver holds the initiative for vehicle operations in a cooperative manner with the assist control. This state occurs when both agents exert torque in the same direction and the vehicle moves in the intended direction.

State II: Driver-led uncooperative state
    The driver holds the initiative for vehicle operations while the DAS attempts to steer against the driver. In this state, the vehicle moves in the driver's intended direction while the DAS exerts torque in the opposite direction.

State III: System-led state, which includes the following two sub-states:

    III-a    System-guided cooperative state
            The human driver is guided by the assist control.

    III-b    System-led uncooperative state
            The human driver resists the assist control.

It should be noted that it is difficult to distinguish between these two sub-states because it requires the measurement of the equivalent torque $\tau_{msl}$, which is generated by muscle force.

State IV: Passive state
    This state rarely occurs in a short time because of inertia or because SAT is dominant.

The present study focuses on the smooth transition from State II to State I. The driver in State II resists the DAS because of intention inconsistency between the driver and system.

Table 1. Cooperative status of the human driver and DAS based on their respective pseudo-work. The cooperative status is judged based on the initiative holder and intent consistency. The right column represents the states in which the driver holds the initiatve, while the upper left cell lists the states in which the DAS holds the initiative. The cooperative status is estimated using $w_c$ and $w_{das}$.

|  |  | $w_c$ | |
|---|---|---|---|
|  |  | $< -\gamma_1^2$ | $\geq -\gamma_1^2$ |
| $w_{das}$ | $\geq -\gamma_2^2$ | (III)-a System-led cooperative ($w_{msl} \geq 0$) <br> (III)-b System-led uncooperative ($w_{msl} < 0$) | (I) Driver-led cooperative |
|  | $< -\gamma_2^2$ | (IV) Passive: No active operation exists | (II) Driver-led uncooperative: driver resisting DAS |

3.3 State Control Methods From State II to State I

Followings are two strategies to change the cooperative state from II to I.

S1) Intent-matching based on estimation of the driver's intent: When the driver's operational intent is estimated using the cooperative status judgment, the cooperative state can be changed from II to I by directly matching the DAS's intent with that of the human driver.

S2) Decreasing the DAS's contribution to vehicle motion: The system's contribution is decreased based on $w_{das}(t)$ to relatively increase the driver's contribution to the vehicle's motion. This is expected to decrease the driver's resistance while waiting for an appropriate change in the DAS's intent by S1.

These ideas are applied to the design of the LKAS in the next section.

## 4. Design of a Lane-Keeping Assist System Based on Cooperative State Estimation

### 4.1 Haptic Guidance Control for Lane-Keeping Assistance

The torque control of Eq. 11 in the domain of Laplace variable $s$, based on a preview driver model (Kondoh, 1958), is employed for haptic lane-keeping control:

$$\tau_{das}(s) = \frac{K(w_{das})}{Ts+1}(L\phi(s)+e(s)), \qquad (11)$$

where $\phi$ is the heading (yaw) angle in the driving lane, $e$ denotes the lateral error in the lane, $K(w_{das})$ (scalar) denotes the gain function, and $T$ is a time constant. Furthermore, $L$ is the preview distance and is given by $L = v\,t_p$, where $v$ is the driving velocity and $t_p$ is a preview time. In this study, $T = 0.15$ s, $t_p = 1.3$ s and $v = 60$ km/h are used throughout the experiments. Using this controller, steering-assist torque is provided to the driver when the vehicle deviates from the center of the lane.

### 4.2 Gain-Tuning Control Method

The controller in Eq. 11 works well when the driver continues to drive in the same lane. However, a mismatch or conflict of intent between the two agents occurs when the human driver decides to change lanes. Such conflicts can be detected by a change in cooperative states, as defined in Section 3. If the driver resists the system and changes lanes, then the cooperative status becomes State II (driver-led uncooperative; see Table 1). Thus, a gain-tuning control method is proposed to achieve a smooth transfer from the original lane to another lane. In the controller, gain $K(w_{das})$ is defined in Eq. 12 as

$$K(w_{das}) := \begin{cases} \dfrac{K_0}{1+\exp(-a\,w_{das}+b)} & \text{(state II)} \\ K_0 & \text{(else)}, \end{cases} \qquad (12)$$

where $a = 10$, $b = 0.4$, and $K_0 = 0.5$ (Fig. 2). Scalar $b$ is determined so that $K$ approaches $K_0$ when $w_{das} = 0$. The scalar $a$ is determined by trial and error so that the gain smoothly reaches its minimum value around the middle of the lane changing.

This control method decreases the gain according to the effort of the system $w_{das}$ that opposes the driver's action in State II. The sigmoid function in Eq. 12 facilitates a smooth shift of gain changes. Thus, strategy S2 of Section 3.3 is implemented.

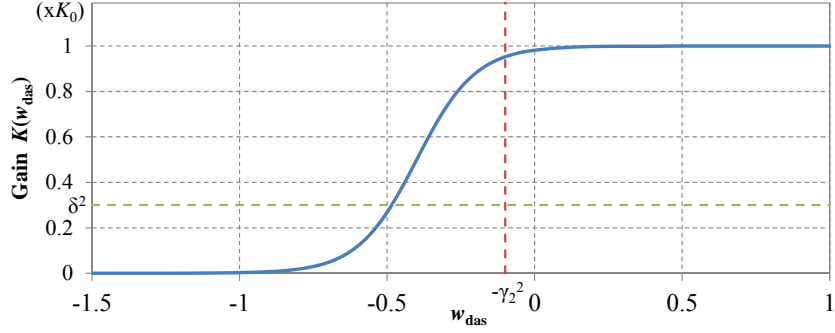

Figure 2. Variable gain for the haptic shared controller of Eq.12. In State II, when $w_{das} < -\gamma_2^2$, the gain $K$ decreased according to $w_{das}$ to decrease the DAS's efforts, which are opposite to the driver's direction. The value of $\delta^2$ on the vertical axis denotes the threshold for detecting the driver's lane-changing intent.

4.3 Algorithm for Changing Lanes

According to strategy S1 in Section 3.3, when the assist system detects the driver's intent to change lanes after being resisted for a specified time, the DAS changes the target lane and smoothly connects the two target lanes, rather than simply weakening the assist control. There are studies on inferring a driver's intent to change lanes based on computational models (Kuge, Yamamura, & Shimoyama, 2000; Salvucci, 2004) and driver's eye-movements (Zhou, Itoh, & Inagaki, 2009) under conditions without any DAS. Tsoi, Mulder, and Abbink (2010) proposed using time-to-line crossing (TLC) to detect a lane change intention and employ a smooth curved line as the new target trajectory, connecting the old and new target lane centers for a haptic LKAS. This method detects a lane-change intention using the vehicle's motion in the lane without considering the conflict between the driver and DAS.

The present paper proposes a method to detect a driver's intent to change lanes using the estimated cooperative state. A method to smoothly connect lane-keeping functions in the old and new target lanes based on the cooperative state is also proposed. A flow chart of the proposed method is given in Fig. 3.

*Detecting Lane-Change Intent*

In State II (driver-led uncooperative), the gain is decreased using our proposed method in Eq. 12. The lane-change intent is defined using the gain $K$ as follows:

$$K(w_{das}) \leq \delta^2 K_0, \qquad (13)$$

where the constant $\delta^2 = 0.3$ was determined by trial and error.

*Changing Lanes*

When the driver's intent to change lanes is detected by Eq. 13, the target lane center, $y_d$, is changed:

$$y_d = \begin{cases} y_{current} - \Delta y & (\dot{y} < 0) \\ y_{current} + \Delta y & (\dot{y} > 0) \\ y_{current} & (else), \end{cases} \qquad (14)$$

where $\Delta y$ is the lane width, and $\Delta y = 3$ m is used in the present paper.

It should be noted that a smooth lane change can be achieved even though this method suddenly changes the target lane, because the gain $K$ gradually decreases during the lane change and then gradually increases in the final part of the lane change, according to Eq. 12.

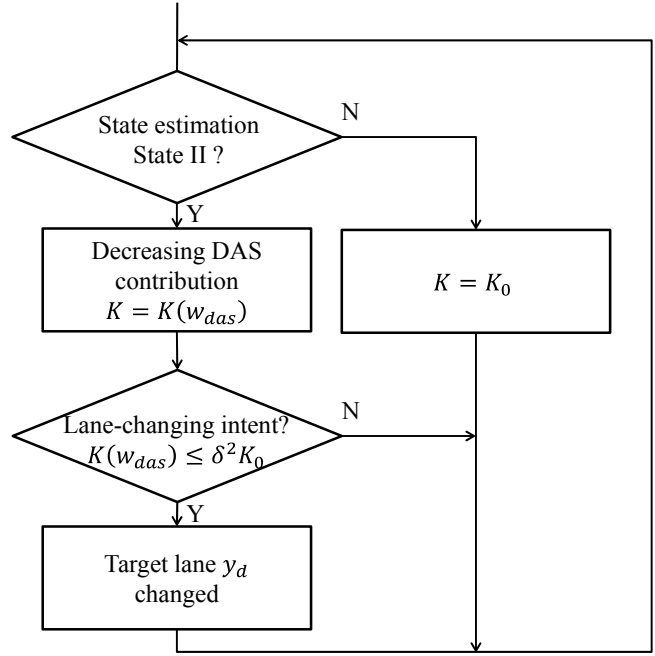

Figure 3. Flow chart of the proposed gain-tuning and lane-change algorithm. Assume that the DAS's lane-keeping function is on. The state estimation function continuously judges the cooperative status. When the state is II, gain-tuning control is activated so that the contribution of the DAS is decreased; otherwise, constant gain is used. In State II, and if lane-changing intent of the driver is detected by the DAS, the target lane is changed.

## 5. User Study

5.1 Apparatus

A fixed-base driving simulator in a desktop configuration was employed for the user study. A 23.6-inch LCD was used for the visual display. The distance between the driver's gaze position and the LCD was 0.9 m. The vehicle dynamics were calculated using CarSim (MSC Corp.), where the E-Class sedan vehicle model with a 3.05 m wheel base was used. A 200 W DC servomotor (RE50, Maxon) with a 1/12 reduction ratio gear head provided the torque generated by the DAS as well as that of the vehicle motion, including the SAT and viscosity. The resultant torques generated by the vehicle and assist system, $\tau_v$ and $\tau_{das}$, respectively, were set by the current control of the DC motor. A six-DOF force-torque sensor (IFS-90M31A50-I50, NITTA) was attached between the steering wheel and one end of the shaft so that the center of its z-axis coincided with the rotational axis to measure torque $\tau_c$.

5.2 Method

*Experimental Scenario*

The test course in the simulator was an endless straight road with 3 m wide lanes. The road had two one-way lanes. The velocity of the host vehicle (HV) was fixed to 60 km/h. The drivers could not control the velocity during the experiments. The participants encountered other vehicles (OVs) in the left lane, which forced them to change lanes to the right and return to the left lane.

There were two scenarios in the experiments (Fig. 4). In Scenario A, three vehicles were travelling at 50 km/h. The center of gravity (CG) distance gap (SAE J2944, 2013) between OVs was 130 m, thus, the participant was forced to overtake one OV in one overtaking. As a result, each participant experienced three overtakings during approximately 1,200 m of driving. In Scenario B, there were eighteen OVs in total; each participant encountered six groups of three vehicles per group. The participant needed to overtake one group of OVs (i.e., three OVs at once), which required more time than passing a single OV (as in Scenario A). The velocities of the vehicles at the end and head of the group were either 30, 40, or 50 km/h. The order of the encountered OV velocities was 40, 30, 50, 40, 30, and 50 km/h. The vehicle in the middle drove just to the left of the HV to prevent the HV from changing back to the left lane. Each participant experienced six overtakings events in Scenario B during approximately 7,500 m of driving. An OV was visible when the CG distance gap between the OV and HV was less than 80 m in both Scenarios A and B.

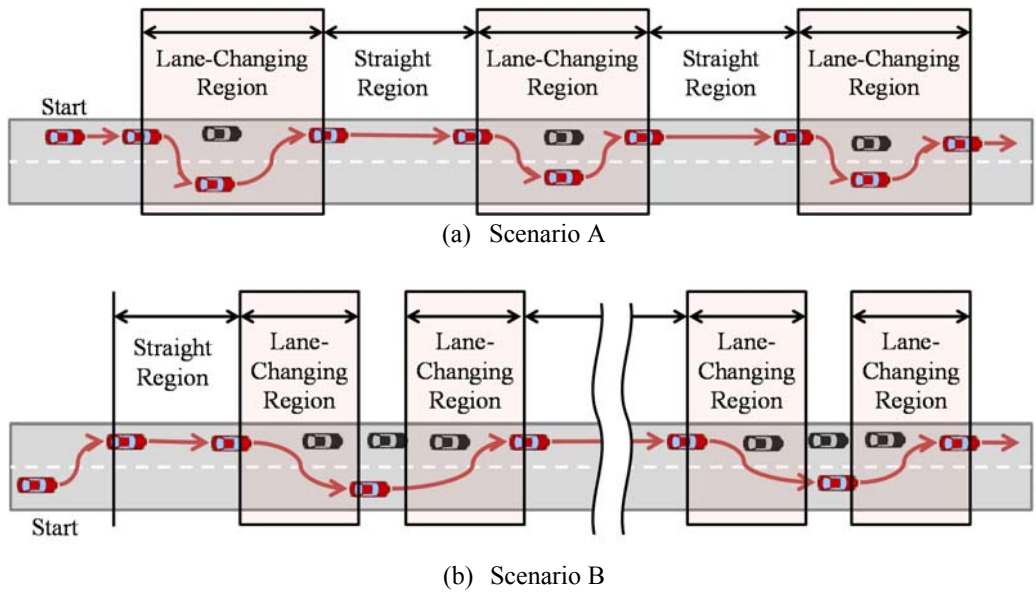

(a) Scenario A

(b) Scenario B

Figure 4. Experimental scenarios. In Scenario A, there are three lane-changing regions with one OV in each region. In Scenario B, there are six groups of three vehicles that are passed by the HV. There are two lane-changing regions for each group: left-to-right and right-to-left. In total, twelve lane-changing regions exist in Scenario B. Straight regions, which are used to analyze lane-keeping performance, exist in the left lane only.

*Experimental Design*

In the user study, the participants were instructed to follow the lead vehicle (LV) in the left of the two lanes travelling in the same direction and overtake the LV when they wanted to overtake it and felt that it was moving slowly enough. The participants were also instructed to return to the left lane after overtaking the LV.

Three levels of system conditions were defined:

a) No system: No DAS was installed in the vehicle.
b) Gain-tuned: The gain-tuning control method (i.e., the proposed system) was activated.
c) TLC: For comparison with the gain-tuned condition, TLC was used to detect the driver's lane-change intention; TLC is defined by

$$TLC := -\frac{y - y_{lm}}{\dot{y}}, \quad (15)$$

where $y$ and $y_{lm}$ denote the lateral position of the HV's CG and lateral position of the lane marker, respectively. When TLC < 1.5 s, the system switches the target lane to the next one. Note that Tsoi et al. (2010) used TLC to change the target lane; however, their method used a smooth trajectory as the target trajectory to connect the target lane centers during the lane change.

*Experimental Procedure*

A within-subject design was employed for each scenario; thus, each participant experienced all three levels of the system conditions. Five males, aged 22 and 23 years, participated in each scenario. The order of the levels was randomized to minimize any order effect.

First, each participant was instructed to drive the test course several times without any OVs to accustom himself to the driving simulator. Participants changed lanes in their own timing. Each participant then experienced three system condition levels in a predetermined order. The experiment of each condition level was composed of practice trials followed by a measurement trial. In the practice trials at each system condition level, each participant drove the test course with at least 10 lane changes in order to become accustomed to driving at the given condition level. In the practice trials, the participants experienced the same scenario as the measurement trial. The participants then drove once at the given condition level in the measurement trial. This procedure was repeated for each condition level.

*Evaluation Method*

*Lateral error*, RMS($e$). The root-mean-square (RMS) value of the lateral error from the lane center in the straight driving area was evaluated by Eq. 16 as a basic performance index of vehicle control for keeping in a lane.

$$RMS(e) := \sqrt{\frac{1}{t_f - t_s} \int_{t_s}^{t_f} e^2(t) dt}, \quad (16)$$

where $e(t)$ denotes the lateral error from the lane center to the HV's CG at each moment $t$. The extracted time period of a straight region is given by $[t_s, t_f]$. A straight driving region is any region in the left lane that excludes lane-changing regions (Fig. 5). A lane-changing region (Fig. 5) begins when steering angle and lateral position changes occur and ends when the lateral position is switched to the next lane and the velocity of the steering angle converges to almost zero. Note that the lane-keeping situation in the right lane is not used for data analysis (Fig. 5).

*Torque generated by driver*, RMS($\tau_c$). The RMS value of the torque exerted on the steering wheel by the driver ($\tau_c$) during lane-changing periods was evaluated by Eq. 17 as an index of driver effort.

$$RMS(\tau_c) := \sqrt{\frac{1}{t_e - t_o} \int_{t_o}^{t_e} \tau_c^2(t) dt}, \quad (17)$$

where the extracted time period is given by $[t_o, t_e]$.

*Maximum torque generated by driver,* $\max(|\tau_c|)$. The peak absolute value of the torque during the lane-changing periods was also evaluated.

*Steering wheel reversal rate (SRR).* The SRR is defined as the number of sign changes in the steering wheel angular velocity per second during the lane-change periods. The SRR was used as an index of driver activity.

*Maximum steering wheel angle*, $\max(|\theta|)$. The peak absolute value of the steering wheel angle during the lane change periods was also evaluated.

## 6. Results

6.1 Examples of HV Response and Gain-Tuning System Performance

Figs. 5 and 6 plot the HV responses for all system conditions and other variables in the gain-tuned condition for Scenarios A and B, respectively. In these figures, time $t = 0$ denotes when the HV's CG crosses a lane marker.

The left and right sides of Fig. 5 show the performance of two different participants. Figs. 5(a) and (b) show little variation in the lateral positions of all system conditions. The cooperative status changes from I to II when the HV starts to move to the right (Figs. 5(e, f)). A few seconds later, the gain, $K(w_{das})$, decreases smoothly in the proposed gain-tuned system (Figs. 5(g, h)); the lane center position, used as the desired lateral position for the gain-tuned system, switches to the right. Subsequently, the cooperative status returns to State I through States IV and III. The gain smoothly increases after the target lane position changes. The same behavior was observed when returning to the left lane.

For the TLC condition, the target lane center has timing similar to that of the gain-tuned system (Figs. 5(c, d)). The DAS torque ($\tau_{das}$) in the gain-tuned condition increases in the opposite direction of the driver's torque just after the driver initiates a lane change. It then decreases by the gain-tuning function and smoothly increases in the same direction as the driver's torque ($\tau_c$) in the latter half of the lane change according to the increased gain (Figs. 5(k, l)). On the other hand, $\tau_{das}$ in the TLC condition decreases rapidly just after increasing in the opposite direction of the driver's torque when the lane change is initiated. Torque in the opposite direction is then generated when the target lane is switched (Figs. 5(k, l)). The driver's torque $\tau_c$ for all system condition levels (Figs. 5(i, j)) varies very little; however, the driver torque $\tau_c$ for no system condition increases slightly at the end of the lane change. For either of the other system conditions in Scenario B, on occasion, $\tau_c$ increases slightly when the DAS torque increases in the opposite direction; this occurs when the driver initiates a lane-change operation, where intent conflict occurs.

The left and right sides of Fig. 6 illustrate the first lane change (to the right) and the second change (to the left), respectively, for Scenario B. Overall, similar signals are measured in both Scenarios A and B.

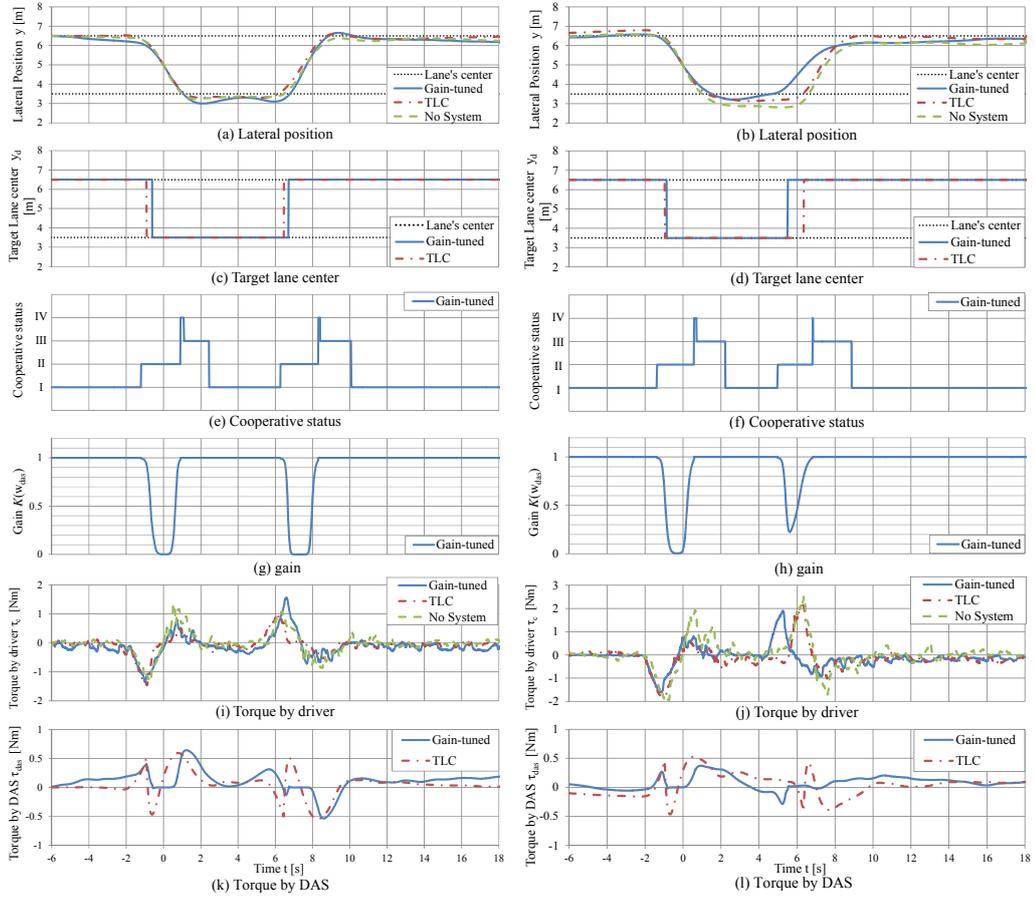

Figure 5. Gain-tuning control system responses in Scenario A: (a) and (b) lateral positions of all system conditions; (c) and (d) target lane center in the DAS; (e) and (f) cooperative status as judged by the DAS; (g) and (h) gain in the shared control; (i) and (j) driver torque; and (k) and (l) DAS torque. The left and right sides show the performance of two different participants.

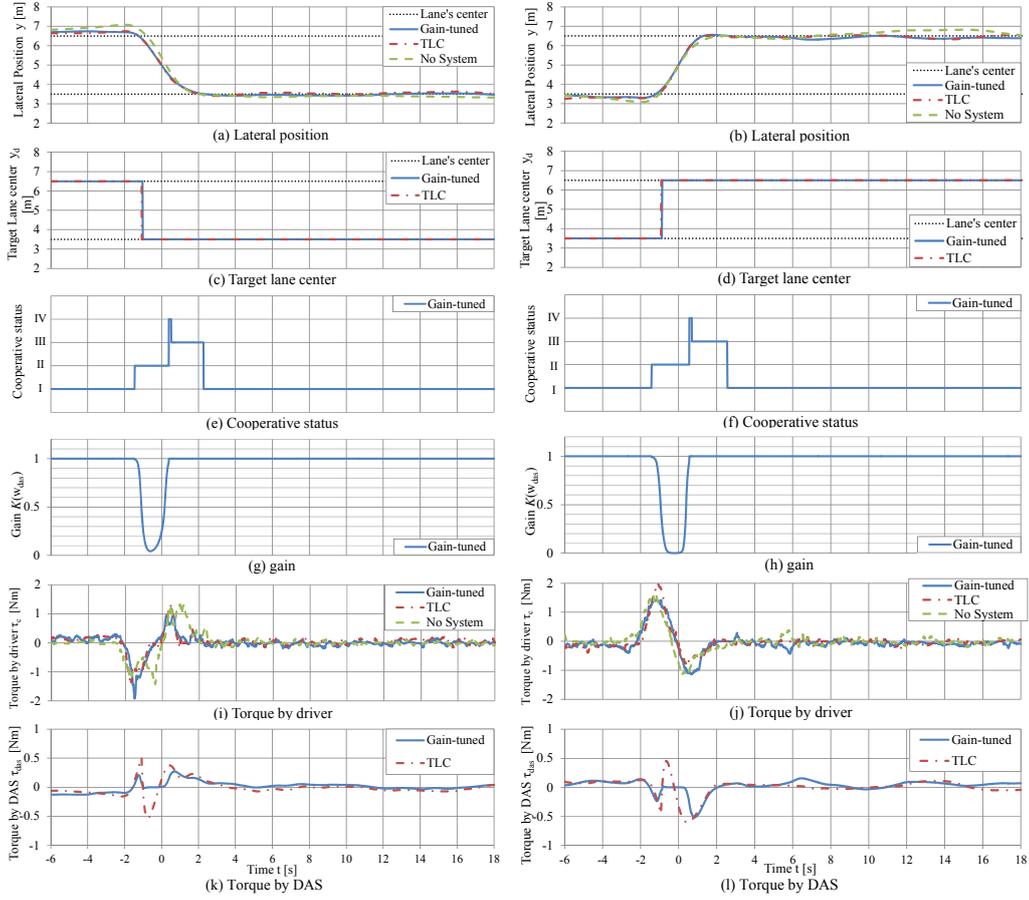

Figure 6. Gain-tuning control system responses in Scenario B: (a) and (b) lateral positions for all system conditions; (c) and (d) target lane center in the DAS; (e) and (f) cooperative status as judged by the DAS; (g) and (h) gain in the shared control; (i) and (j) driver torque; (k) and (l) DAS torque. The left and right sides showed the performance of two different participants.

6.2 RMS($e$): Lateral Error

Fig. 7 shows the RMS($e$) of the lateral error when the HV drives in a straight region. The mean and standard deviations of the lateral errors are shown in Table 2. In both Scenarios A and B, these results suggest that the mean lateral error for either system, gain-tuned or TLC, is smaller than that of the no system condition. In Scenario A, the lateral error for the TLC system is smaller than that of the gain-tuned system. These results demonstrate that both assist systems decrease the lateral error in the lane-keeping phase.

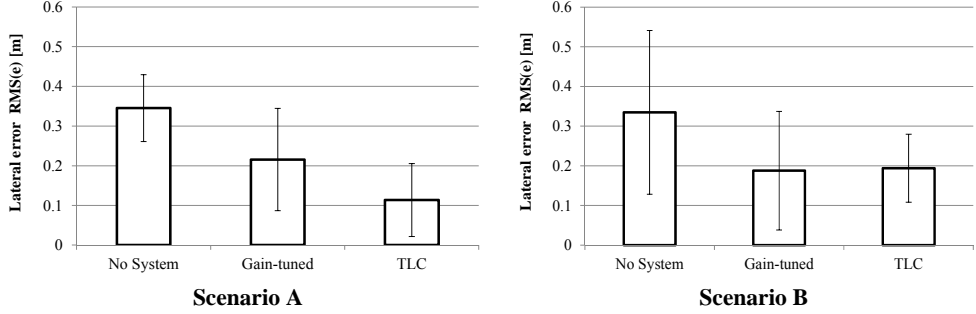

Figure 7. RMS($e$): RMS of the lateral error in straight driving regions. Bars indicate the mean of all participants. Error bars indicate the standard deviation.

### 6.3 RMS($\tau_c$): Driver Torque

Fig. 8 shows the RMS values of the driver torque, $\tau_c$, during lane changes. The mean and standard deviation are shown in Table 2. The results for Scenario A suggest that there is no differences between the RMS($\tau_s$) of the three system conditions. In Scenario B, the results suggest that there is no difference between the torque for the no system condition and the TLC condition; however, the torque of the gain-tuned system is slightly larger than that of the other conditions.

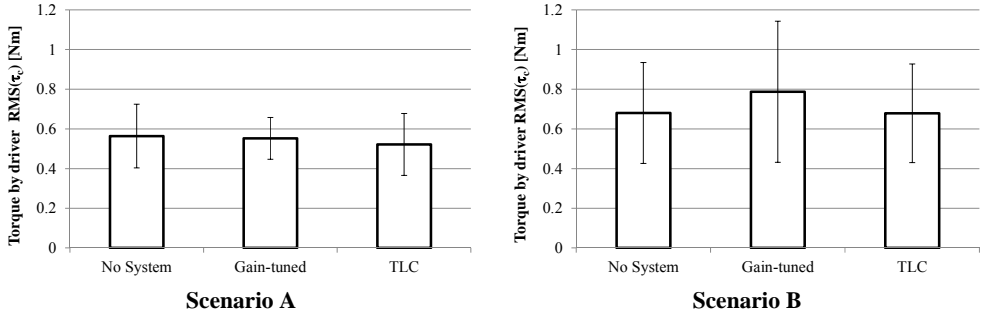

Figure 8. RMS($\tau_c$): RMS of the torque generated by the driver in lane change regions. Bars indicate the mean of all participants. Error bars indicate the standard deviation.

### 6.4 Max($|\tau_c|$): Peak Absolute Value of Driver Torque

Fig. 9 presents the maximum torque generated by the driver. The mean and standard deviation are shown in Table 2. The results in Scenario A suggest that there are no differences between the three system conditions. In Scenario B, the results suggest that there is no difference between the gain-tuned and TLC systems; whereas the maximum driver torque for no system conditions could be smaller than that of the other systems.

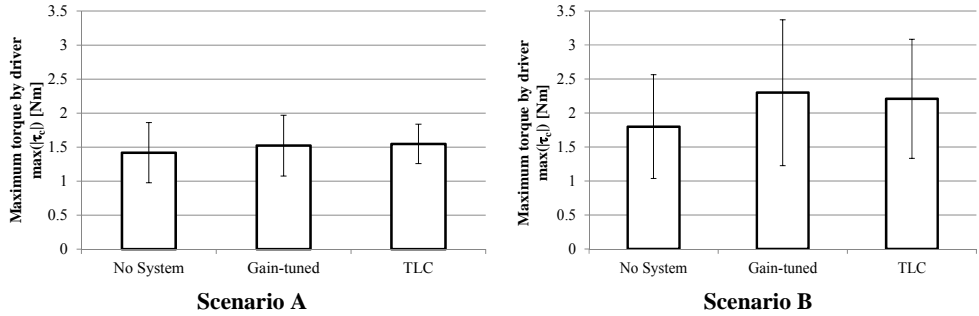

Figure 9. Max($|\tau_c|$): Peak absolute value of the torque generated by the driver during in lane changing regions. Bars indicate the mean of all participants. Error bars indicate the standard deviation.

6.5 SRR

SRR values for the lane-change regions are shown in Fig. 10. The mean and standard deviation are listed in Table 2. The results for Scenario B suggest that there are no differences between the three system conditions. For Scenario A, the results suggest that the SRR for no system conditions is slightly larger than that of the others; however, this difference is not physically significant.

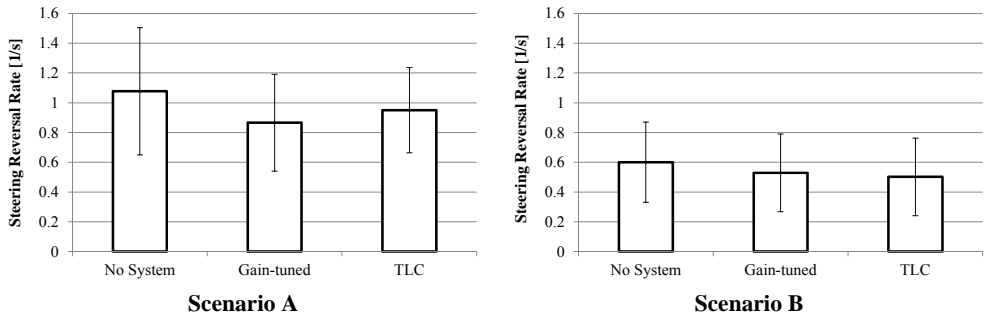

Figure 10. SRR in lane change regions. Bars indicate the mean of all participants. Error bars indicate the standard deviation.

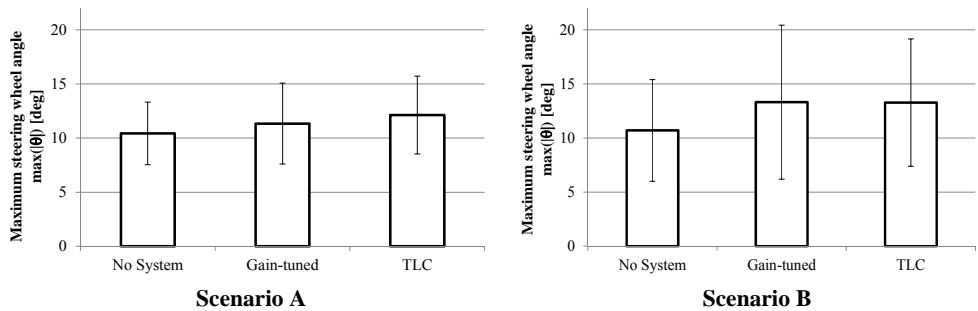

Figure 11. Max($|\theta|$): Peak absolute value of the steering wheel angle.

Table 2. Mean and standard deviation of evaluated values.

|  | Scenario A | | | Scenario B | | |
| --- | --- | --- | --- | --- | --- | --- |
|  | No system | Gain-tuned | TLC | No system | Gain-tuned | TLC |
| RMS(e) [m] | 0.345(0.084) | 0.216(0.129) | 0.114(0.092) | 0.335(0.206) | 0.188(0.150) | 0.194(0.086) |
| RMS($\tau_c$) [Nm] | 0.564(0.161) | 0.552(0.105) | 0.522(0.156) | 0.680(0.254) | 0.787(0.355) | 0.679(0.249) |
| max($|\tau_c|$) [Nm] | 1.42(0.422) | 1.52(0.447) | 1.55(0.290) | 1.80(0.763) | 2.30(1.07) | 2.21(0.876) |
| SRR [1/s] | 1.08(0.427) | 0.866(0.325) | 0.950(0.287) | 0.601(0.270) | 0.529(0.261) | 0.502(0.261) |
| max$|\theta|$ [deg] | 10.43(2.89) | 11.3(3.75) | 12.1(3.60) | 10.7(4.69) | 13.3(7.12) | 13.3(5.88) |

6.6 Max($|\theta|$): Peak Absolute Value of Steering Wheel Angle

Fig. 11 shows the peak absolute values of the steering wheel angle. The mean and standard deviation are listed in Table 2. The results in Scenario A suggest that there are no differences between the three system conditions. In Scenario B, the results suggest that there is no difference between the gain-tuned and TLC systems, while the angle for no system conditions could be smaller than that of the others.

## 7. Discussion

As shown in Figs. 5 and 6, the proposed method successfully judged the cooperative status using the pseudo-work, changed the gain of the control, and matched the DAS's intent with the driver's. Consequently, the resultant trajectory of the vehicle's lateral displacement during lane changes did not greatly differ from that without any LKAS.

As shown in Fig. 7, the RMS($e$) of the DAS, gain-tuned, and TLC conditions were smaller than that without the DAS. This demonstrates the effectiveness of the basic lane-keeping assist functions of the proposed DAS. These results agree with the results of previous studies (e.g., Abbink & Mulder, 2009; Tsoi et al., 2010).

Next, we discuss the stability and activity variables during lane changes. There were no significant differences between the RMS($\tau_c$) and max($\tau_c$) for the system conditions in Scenario A, which could be regarded as a double lane change. In Scenario B, which can be regarded as two separate lane changes, the results suggest that the RMS($\tau_c$) and max($\tau_c$) were larger for two conditions with the DAS than that for no system condition. These results suggest that the torque, $\tau_c$, increases at the beginning of the lane change with the DAS even though it is not as large. The fact that RMS($\tau_c$) for the TLC condition in Scenario B was smaller than that of the gain-tuned condition implies that the TLC method assists the driver's torque after switching the target lane. From the time course of the torque $\tau_c$ and $\tau_{das}$, there were cases that $\tau_c$ increased slightly when the driver's lane-change initiation caused $\tau_{das}$ to increase; this change was in the opposite direction of $\tau_c$ because a conflict of intent occurred between the driver and DAS. However, gain-tuned control and target lane switching successfully decreased the opposing torques. The gain-tuned control also facilitated smooth changes in $\tau_{das}$, while $\tau_{das}$ of the TLC condition changed sharply.

The SRR was used as the index of the driver's activity or stability of the steering operation. There was no large difference in the SRR between the system conditions in both Scenarios A and B. In addition, there were no significant differences in max ($|\theta|$) in Scenario A, while that of the DAS condition were larger in Scenario B. However, as the difference in the average steering wheel angle was within three degrees; this difference is not significant physically. These results demonstrate that the proposed method can change lanes with activity and smoothness similar to manually driving without a DAS.

The above discussion suggests that the proposed method performs well on straight roads and follows a driver's initiated lane changes as smoothly as manual driving, by introducing the judgment of cooperative status, gain-tuning control, and switching target lanes. It should be noted that the system with the TLC condition works almost identically to the proposed method; however, in the TLC method, a sharp torque change occurred. It is problematic if a driver decides not to change lanes because it affects the sudden target lane change in the TLC condition. With the proposed method, this is not an issue because the gain is decreased during lane changes and smoother torque changes are achieved.

One of the limitations of the present study is seen in a difficultly to generalize deductions here, since our results were obtained from a user study with five participants in each scenario. However, the present study demonstrates the existence of users who are able to effectively use the proposed method. In the present study, the lane-keeping function in the straight road was investigated. It is thought that the proposed method can be applied to curved roads by adding curvature compensation to the preview driver model.

## 8. Conclusion

A new method to estimate the cooperative status between a driver and DAS in the haptic shared control of a steering-assist system was proposed from the viewpoints of initiative holder and intention consistency. These factors were based on the pseudo-work exerted on the vehicle motion by the steering input of a driver or a DAS control actuator. A gain-tuning control method was proposed for lane-keeping assistance to enable the driver to change lanes smoothly.

The driving simulator experiments demonstrated that the proposed method can appropriately estimate the cooperative status. The proposed gain-tuning control for the lane-keeping assist facilitated smooth lane changes. The proposed system and the TLC system improved lane-keeping performance in straight driving with haptic guidance; furthermore, these systems enabled smooth driver-initiated lane changes without significantly impacting the driver's torque or steering angle. Moreover, the proposed system allowed for smoother changes in the driver's torque during lane changes, unlike the TLC system.

Finally, the present study demonstrated that the proposed method can be used effectively. In fact, results were successfully obtained from five participants for each scenario. Increasing the number of participants is an important direction for future work to generalize the results. In addition, the present study did not deal with sensor noise, which causes erroneous estimation of the cooperative status. A user test including such erroneous conditions is another issue for future work. Expansion to various road geometries such as curved roads should be done, since the present study only considered straight roads. In addition, the proposed method of cooperative states judging will be applied to other situations, such as control from the system-led state to the driver-led state.

## Acknowledgments

The authors would like to thank the anonymous reviewers for their insightful comments and suggestions. This work was partially supported by the JSPS KAKENHI Grant-in-Aid for Scientific Research (A), Grant Number 26242029.

Authors' names and contact information:
R. Nishimura, Ritsumeikan University, Japan, Email: is0032pv@ed.ritsumei.ac.jp;
T. Wada, Ritsumeikan University, Japan, Email: twada@fc.ritsumei.ac.jp;
S. Sugiyama, Ritsumeikan University, Japan, Email: seijisan@hr.ci.ritsumei.ac.jp.